\documentclass[sigconf,nonacm]{acmart}
\usepackage{amsmath}
\usepackage{graphicx}
\usepackage{booktabs}
\usepackage{multirow} 
\usepackage{algorithm}
\usepackage{algorithmic}
\usepackage{makecell}
\usepackage{float}
\begin{document}

\title{Efficient Training with Foresight: Multi-Token Auxiliary
Supervision for Autoregressive Image Generation}

\author{Guo Niu}
\authornote{The first two authors contributed equally to this work.}
\orcid{0000-0002-1552-7310}
\affiliation{%
  \institution{Foshan University}
  \city{Foshan}
  \country{China}}
\email{niuguo@fosu.edu.cn}

\author{Xiongfei Yao}
\authornotemark[1]
\orcid{0009-0009-8464-8448}
\affiliation{%
  \institution{Foshan University}
  \city{Foshan}
  \country{China}}
\email{2112453049@stu.fosu.edu.cn}

\author{Teng Wang}
\correspondingauthor
\orcid{0000-0003-2331-3619}
\affiliation{%
  \institution{The University of Hong Kong}
  \city{Hong Kong}
  \country{China}}
\email{tengwang@connect.hku.hk}

\author{Nannan Zhu}
\orcid{0000-0003-4038-3053}
\affiliation{%
  \institution{Sun Yat-sen University}
  \city{Guangzhou}
  \country{China}}
\email{zhunn25@mail.sysu.edu.cn}


\begin{abstract}
  Autoregressive (AR) image generation has shown strong potential for scalable high-fidelity synthesis by modeling images as discrete token sequences. However, traditional next token prediction (NTP) continues to suffer from  sparse and myopic supervision, insufficiently discriminative representations, and high training cost caused by dense computation over the full token sequence. To address these issues, we propose multi-token autoregressive (MTAR), a unified training framework that improves autoregressive image generation from three aspects: prediction objectives, representation regularization, and training efficiency. Specifically, MTAR introduces multi-token prediction (MTP) to alleviate the sparsity and myopia of traditional NTP by imposing joint supervision on multiple future tokens; employs token-level contrastive regularization (TCR) to explicitly enhance the separability of sampled token representations and thereby improve representation discriminability; and incorporates semantic dropping (SD) as a semantics-aware training acceleration strategy to reduce redundant computation on low-information tokens while preserving informative learning signals. All three components are applied only during training and introduce no additional overhead during autoregressive inference. On ImageNet, MTAR achieves a better balance between generation quality and training efficiency. Compared with LlamaGen, MTAR achieves up to 0.95 lower FID and 39\% faster training. Moreover, even with only 1/3 of the training iterations, it still attains performance comparable to or better than the baseline, substantially reducing training time. Code is available at https://github.com/yzh595/MTAR-Code.
\end{abstract}




\maketitle

\section{Introduction}
In recent years, AR image generation~\cite{radford2018improving,pmlr-v48-oord16,touvron2023llama,liu2024customize,ren2025beyond,Pang_2025_CVPR,salimans2016improved,vqvae_nips,wang2024emu3,NEURIPS2024_e91bf7df,yuan2025argrpo,yu2026frequencyautoregressiveimagegeneration,huang2025nfigmultiscaleautoregressiveimage,yi2025iar2improvingautoregressivevisual} has demonstrated great potential in both scalability and synthesis quality by discretizing images into token sequences and modeling them token by token under the maximum likelihood objective. However, the classical autoregressive training paradigm typically flattens a two-dimensional image into a one-dimensional sequence and relies solely on NTP as the supervision signal, that is, predicting the next token conditioned on the generated prefix.

Despite its simplicity and effectiveness, this paradigm still suffers from several inherent limitations. First, single-step prediction provides supervision for only one position at each step, resulting in sparse training signals. Second, the conventional NTP objective imposes no explicit constraint on the geometric structure of hidden token representations, which may restrict their diversity and discriminability in the feature space and consequently increase the risk of degenerate phenomena, such as repetitive textures and representation collapse. Moreover, autoregressive image generation requires dense computation over the full token sequence during training, although the semantic importance of different tokens is highly uneven. Consequently, substantial computation is inevitably allocated to low-information tokens, leading to redundant overhead and reduced training efficiency.

\begin{figure}[t]
\centering
\includegraphics[width=1\linewidth]{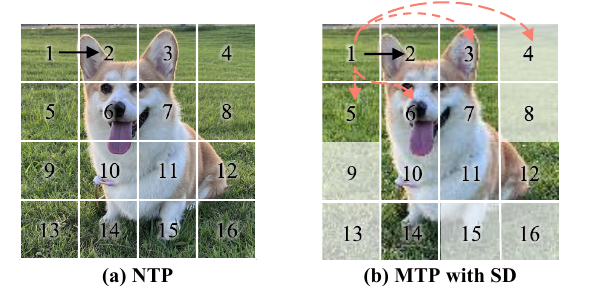}
\caption{(a) Traditional NTP provides only short-range supervision over the full token sequence, leading to sparse and myopic training signals, while requiring autoregressive modeling over the entire sequence with high computational cost. (b) Our framework improves training with MTP and SD. MTP introduces more forward-looking and denser auxiliary supervision signals, thus enhancing contextual dependency modeling and alleviating the myopic nature of conventional single-step prediction, as indicated by the orange-red dashed arrows. SD further improves training efficiency by discarding less informative patches; the white patches denote the dropped regions.}
\Description{A two-panel schematic comparing conventional next-token
prediction with the proposed training framework. Panel (a) shows all image
tokens retained and connected primarily through short-range, single-step
prediction signals. Panel (b) shows retained tokens receiving additional
orange-red dashed connections to more distant future targets, representing
multi-token prediction. Several image patches are colored white to indicate
tokens removed by semantic dropping, resulting in a shorter training
sequence.}
\label{fig:Semantic Dropping and Multi-Token Prediction}
\end{figure}

To address these limitations, we improve autoregressive image generation in terms of prediction supervision, representation regularization, and training efficiency. We introduce  MTP as an auxiliary training objective to alleviate the sparse supervision and inherent myopia of single-step prediction. By imposing joint supervision on multiple future tokens, MTP better exploits the intrinsic two-dimensional structure of images, as illustrated in Fig~\ref{fig:Semantic Dropping and Multi-Token Prediction}(b). This design provides denser training signals for the shared contextual representation in a manner more consistent with image geometry, allowing the model to better capture local structural patterns and future contextual information.
To further compensate for the lack of explicit constraints on hidden token features in the traditional NTP objective, we introduce TCR. TCR treats other sampled tokens as negative samples and explicitly enhances the separability among different token representations through cross-view contrast, which reduces representation redundancy and improves the discriminability of hidden representations.

In addition, we propose SD as a training acceleration strategy motivated by the observation that low-information tokens still consume substantial computation in traditional AR training. Based on patch-wise importance scores extracted offline by an external vision encoder, SD performs semantics-aware sampling and preferentially retains semantically salient patches during training, which reduces training cost and improves training efficiency while preserving informative learning signals as much as possible.

\begin{figure}[t]
\centering
\includegraphics[width=1.0\linewidth]{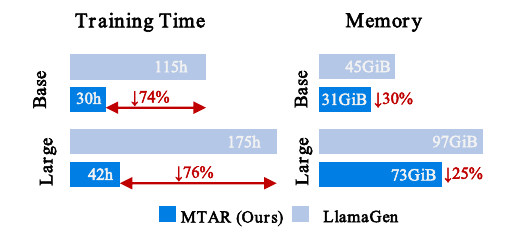}
\caption{Comparison of total training time (hours) and total GPU memory usage across all training GPUs between LlamaGen~\cite{sun2024autoregressive} and MTAR for the Base and Large models, along with the corresponding FID values. LlamaGen is trained for 300 epochs, whereas MTAR is trained for only 100 epochs. For the efficiency measurements, the Base models are trained on a single NVIDIA A800 GPU, while the Large models are trained on two NVIDIA A800 GPUs. MTAR consistently achieves better generation quality and efficiency than the corresponding LlamaGen baselines: for the Base model, FID improves from 5.46 to 5.37, while training time and memory are reduced from 115 to 30 and from 45\,GiB to 31\,GiB; for the Large model, FID improves from 3.80 to 3.36, while training time and memory are reduced from 175 to 42 and from 97\,GiB to 73\,GiB. Both methods use \texttt{torch.compile}.}
\Description{A grouped comparison of LlamaGen and MTAR across the Base and
Large model settings, showing training time, GPU memory usage, and FID.
MTAR has lower values than LlamaGen for all three metrics in both settings.}
\label{fig:100epoch}
\end{figure}

We conduct extensive experiments on the ImageNet image generation benchmark~\cite{deng2009imagenet}, which thoroughly validate the significant advantages of MTAR in both generation quality and training efficiency. Under the same number of training iterations, MTAR achieves substantially better generation performance than LlamaGen~\cite{sun2024autoregressive}, improving the FID from 3.80 to 2.85 while also making training 39\% faster. Furthermore, as shown in Fig~\ref{fig:100epoch}, even with only 1/3 of the training iterations, MTAR still attains performance comparable to or even better than the LlamaGen~\cite{sun2024autoregressive} baseline, demonstrating that the proposed method significantly improves training efficiency while enhancing generation quality.

Our main contributions are summarized as follows:
\begin{itemize}

\item We propose MTAR, a unified training framework for autoregressive image generation, which systematically improves the conventional autoregressive training paradigm from three aspects: prediction supervision, representation regularization, and training efficiency.
\item We introduce MTP and TCR to enhance autoregressive training from the perspectives of supervision signal enrichment and hidden representation regularization, respectively, improving the model’s contextual modeling ability and the discriminability of token representations.
\item We propose SD as a semantics-aware training acceleration strategy, which reduces redundant computation by preferentially retaining more informative tokens, while improving training efficiency and preserving effective learning signals.

\end{itemize}

\begin{figure*}[t]
\centering
\includegraphics[width=1\linewidth]{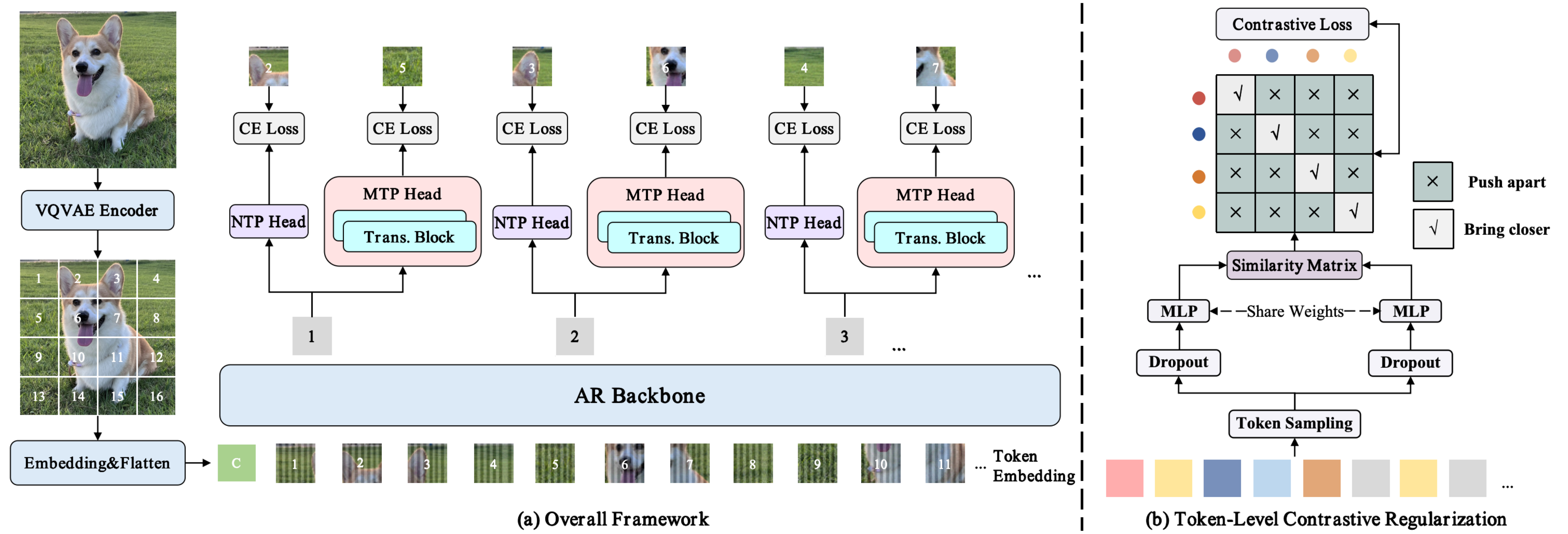}
  \caption{\textbf{Overview of the proposed framework.}
  (a) \textbf{Overall framework under the MTP (R$_1$, B) setting:} besides the original NTP (R$_1$) head, an additional MTP head is introduced to provide auxiliary supervision for the token directly below the current token.
(b) \textbf{TCR:}
Sampled token features are passed through two Dropout-perturbed branches and a shared MLP to construct a cross-view similarity matrix. The diagonal entries bring matched token pairs closer, while the off-diagonal entries push different sampled tokens apart.}
\Description{A two-panel overview of MTAR. Panel (a) shows the NTP head
predicting the next token and an MTP head predicting the token below.
Panel (b) shows TCR constructing a similarity matrix from two
Dropout-perturbed views of sampled token features.}
\label{fig:overall_framework}
\end{figure*}

\section{Related Work}

\subsection{Autoregressive Visual Generation}

Autoregressive visual generation models typically follow a GPT-style next-token prediction paradigm~\cite{radford2019language,achiam2023gpt}, where each token is generated conditioned on all previous tokens~\cite{vqgan,ldm,parti,pang2024next,he2024zipar,luo2024open,NIPS2016_b1301141,pmlr-v119-chen20s,chang2023muse,xie2024show}. A representative example is LlamaGen~\cite{sun2024autoregressive}, which formulates image synthesis as token-by-token prediction in raster order with a simple and efficient framework. Beyond this paradigm, MaskGIT~\cite{chang2022maskgit} and MAR~\cite{li2024autoregressive} support parallel token prediction during training through bidirectional attention and masked prediction. PAR~\cite{Wang_2025_CVPR} accelerates sampling by grouping tokens, while VAR~\cite{var} improves visual fidelity through multi-scale prediction at the cost of higher computation. NAR~\cite{he2025neighboring} focuses on local dependencies between adjacent tokens and adopts parallel decoding, which changes the inference process and limits its generality. RAR~\cite{Yu_2025_ICCV} further improves autoregressive training through randomized token ordering while preserving the standard autoregressive framework. However, existing methods still struggle to jointly address sparse supervision, weak discriminability of hidden representations, and high training cost. To address these limitations, we propose MTAR, which enhances the classical autoregressive training paradigm from three aspects: prediction objectives, representation regularization, and training efficiency. Without changing the inference mechanism, MTAR improves both generation quality and training efficiency.

\subsection{Multi-Token Prediction}

Multi-token prediction  has been widely studied in language modeling~\cite{gloeckle2024better,qi2020prophetnet,NEURIPS2018_c4127b91,cai2024medusa,liu2024deepseek,liu2025mtp,zhang2025enhancing,samragh2025your}. Related methods typically strengthen training by jointly predicting multiple future tokens and, in some cases, support parallel decoding to improve inference efficiency. Gloeckle~\cite{gloeckle2024better} shows that the benefits of MTP become more pronounced at larger model scales, while also enabling inference acceleration through parallel multi-token decoding. DeepSeek-V3~\cite{liu2024deepseek} also incorporates MTP into large-scale language model training and highlights its potential for inference acceleration via speculative decoding. However, existing MTP methods are mainly developed for language modeling, where prediction targets are defined over one-dimensional sequential dependencies. In autoregressive visual generation, although images are flattened into one-dimensional token sequences during training, the tokens inherently originate from a two-dimensional spatial grid. Therefore, applying MTP to vision is not a direct transfer of language-modeling designs; auxiliary supervision targets should instead account for the spatial locality and two-dimensional neighborhood structure of images, rather than relying only on linear offsets in raster order.
Based on this distinction, we introduce MTP into autoregressive visual generation as a training-time auxiliary supervision mechanism, rather than focusing on its parallel decoding capability.  MTAR defines auxiliary prediction targets from two-dimensional spatial relations, providing denser and more geometry-aware supervision for the shared contextual representation. This improves the modeling of local structures and future contextual information while preserving the standard autoregressive inference pipeline.

\section{MTAR Framework}

\subsection{Multi-Token Prediction}

We introduce a MTP strategy on top of the conventional  NTP paradigm. MTP uses one or more auxiliary heads to impose additional supervision on future tokens at different offsets, providing denser training signals for shared contextual representations in a way that better aligns with image geometry and mitigates the myopic nature of single-step prediction, as illustrated in Fig~\ref{fig:overall_framework}. This design encourages the model to capture richer local structures and future contextual information.

Let $\{y_1, y_2, \dots, y_N\}$ denote the sequence of $N$ image tokens. Given the condition token and the image tokens, the backbone produces hidden features $H=\{h_0,h_1,\dots,h_{N-1}\}$, where $h_0$ corresponds to the condition token and each $h_t$ is used to predict $y_{t+1}$.

\noindent\textbf{NTP Head.}
The NTP head is implemented as a linear output layer that maps $h_t$ to the conditional distribution of the next token $y_{t+1}$. The NTP loss is defined as
\begin{equation}
\mathcal{L}_{\text{NTP}} = \sum_{t=0}^{N-1} -\log P(y_{t+1} \mid h_t),
\label{eq:ntp_loss}
\end{equation}
During inference, the auxiliary MTP heads are discarded, and only the NTP head is used for traditional autoregressive decoding.

\noindent\textbf{MTP Head.}
Unlike the simple NTP head, each auxiliary MTP head consists of two additional Transformer blocks followed by a linear output layer. Specifically, for the $w$-th auxiliary MTP head, the backbone feature $h_t$ is further transformed into a head-specific enhanced feature $h_{t,w}^{\text{en}}$, which is used to predict the target token at an additional offset $\delta_w$ in raster order. Here, $\delta_w$ denotes the prediction offset associated with the $w$-th auxiliary MTP head, namely the relative distance from the current position to the target token. For example, $R_2$ corresponds to $\delta=2$, while $B$ corresponds to a vertical offset equal to the token-grid width\ $G$. The $R_1$ prediction (i.e., $\delta=1$) is already handled by the NTP head and is therefore not counted as an auxiliary MTP head. In the default MTP $(R_1, B)$ setting, the model uses the original NTP head to perform the $R_1$ prediction and introduces one additional auxiliary MTP head with offset $\delta=G$, where $G$ denotes the width of the token grid (e.g., $G=16$ for a $16\times16$ token grid ).

The overall MTP objective aggregates the auxiliary supervision provided by all auxiliary MTP heads:
\begin{equation}
\mathcal{L}_{\text{MTP}} =
\sum_{w=1}^{W}
\sum_{t=0}^{N-1-\delta_w}
-\log P_w \left( y_{t+\delta_w} \mid h_{t,w}^{\text{en}} \right),
\label{eq:mtp_loss}
\end{equation}
where $W$ denotes the number of auxiliary MTP heads (excluding the NTP head), $h_{t,w}^{\text{en}}$ denotes the enhanced feature produced by the $w$-th auxiliary MTP head at position $t$, and $P_w(\cdot)$ denotes the predictive distribution of the corresponding auxiliary head.

The effectiveness of MTP depends on the design of the auxiliary MTP heads, including their number and the associated additional prediction directions. In practice, these directions are implemented as different offsets in raster order relative to NTP.

\begin{figure*}[t]
\centering
\includegraphics[width=1\linewidth]{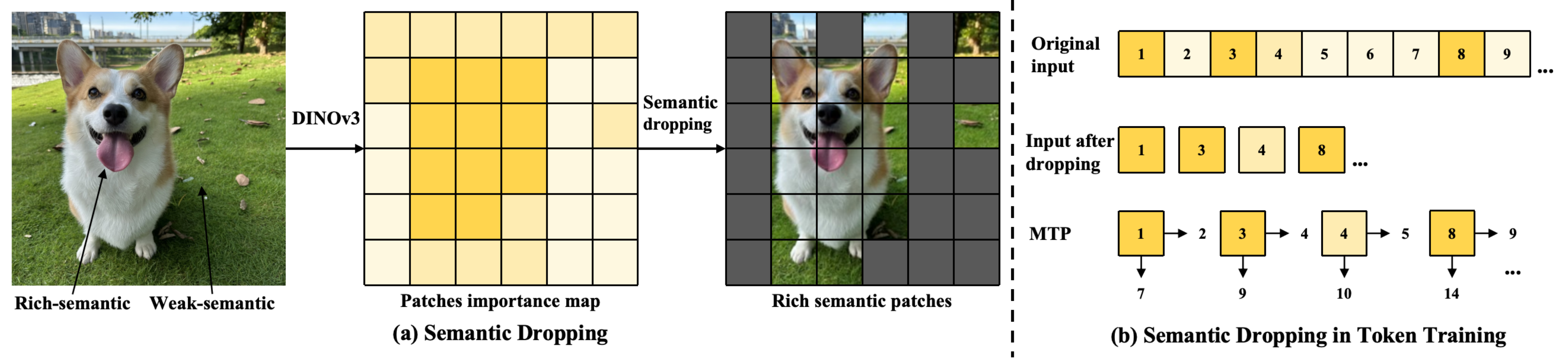}

  \caption{\textbf{Semantic Drop.}
(a) DINOv3~\cite{simeoni2025dinov3} extracts a token importance map for each image 
(darker: higher semantic importance). 
Low-importance tokens are dropped, retaining only semantically 
salient tokens for training. 
(b) Retained tokens preserve their original spatial indices, 
and both NTP and MTP losses are computed accordingly.
  }
  \Description{DINOv3 identifies important image tokens, which retain their
original positions for NTP and MTP training.}
  \label{fig:Semantic Drop}

\end{figure*}

\subsection{Token-Level Contrastive Regularization}

The traditional NTP objective directly supervises output predictions, but does not explicitly constrain hidden token representations. As a result, the learned token representations may be insufficiently discriminative for long-sequence autoregressive modeling. To address this issue, we introduce TCR, as illustrated in Fig~\ref{fig:overall_framework}(b), to impose lightweight regularization on hidden token representations during training.

Specifically, after removing the condition token, we randomly sample $K$ features from the $B \times N$ image-token features in a batch to form a feature matrix $\mathbf{F}\in\mathbb{R}^{K\times D}$, where $B$ denotes the batch size, $N$ denotes the number of image tokens, and $D$ denotes the hidden feature dimension. We then apply two independently sampled Dropout perturbations to $\mathbf{F}$ to obtain two random views:
$\tilde{\mathbf{F}}_1=\mathcal{D}(\mathbf{F};m_1),
\tilde{\mathbf{F}}_2=\mathcal{D}(\mathbf{F};m_2),$
where $\mathcal{D}(\cdot;m)$ denotes the Dropout perturbation parameterized by mask $m$, and $m_1$ and $m_2$ are two independent Dropout masks. The two views are further fed into a shared MLP projection head followed by $\ell_2$ normalization, yielding
$\mathbf{Z}_1=\mathrm{norm}\big(f_\theta(\tilde{\mathbf{F}}_1)\big),
\mathbf{Z}_2=\mathrm{norm}\big(f_\theta(\tilde{\mathbf{F}}_2)\big).$

Based on these two views, we compute a cross-view similarity matrix
$\mathbf{S}_{ij}=\frac{\mathbf{z}_{1,i}^{\top}\mathbf{z}_{2,j}}{\tau},$
where $\tau$ is a temperature parameter. The diagonal term $\mathbf{S}_{ii}$ measures the matching score of the same token under the two perturbed views, while the off-diagonal terms measure its similarity to other sampled tokens. TCR is optimized with an InfoNCE-style~\cite{pmlr-v119-chen20j,He_2020_CVPR} contrastive loss in cross-entropy form:
\begin{equation}
\mathcal{L}_{\mathrm{TCR}}
=-\frac{1}{K}\sum_{i=1}^{K}
\log
\frac{\exp(\mathbf{S}_{ii})}
{\sum_{j=1}^{K}\exp(\mathbf{S}_{ij})}.
\label{eq:tcr_loss}
\end{equation}
We adopt a one-sided InfoNCE formulation, where tokens from $\mathbf{Z}_1$ serve as queries against $\mathbf{Z}_2$. The matched token across the two views forms the positive pair, while the remaining sampled tokens are treated as negatives.

Unlike ST-AR~\cite{yue2025understand}, which relies on an EMA teacher, data augmentation, and inter-step/inter-view contrastive objectives, TCR directly constructs token-level positive pairs from Dropout perturbations and applies contrastive regularization to the sampled hidden features in the current batch. As with other contrastive formulations based on in-batch negatives, a small number of false negatives may exist in principle. 
Using DINOv3~\cite{simeoni2025dinov3} features on 10,000 ImageNet training images, only 2.01\% and 8.73\% of token pairs have cosine similarity above 0.9 and 0.8, respectively, suggesting that false negatives have limited practical impact under TCR sampling.

\subsection{Semantic Dropping}

Traditional autoregressive training does not distinguish the semantic importance of different tokens, causing a substantial portion of training computation to be allocated to low-information patches. To overcome this limitation, we propose SD, as illustrated in Fig~\ref{fig:Semantic Drop}, which prioritizes semantically salient patches during training to improve training efficiency.

For each training image, we extract patch-level semantic importance scores $\mathbf{s}\in\mathbb{R}^{N}$ offline using an external vision encoder, and retain $M=\lfloor N(1-r)\rfloor$ patches given a drop rate $r\in(0,1)$. Specifically, we sample without replacement according to the temperature-scaled distribution
$p_i=\frac{s_i^{1/\tau^{\prime}}}{\sum_j s_j^{1/\tau^{\prime}}},$
to obtain the retained set $\mathcal{I}_{\text{keep}}$, where $\tau^{\prime}$ controls the concentration of the sampling distribution. The retained indices are sorted according to their original spatial order before being fed into the backbone. Each retained token is assigned a 2D RoPE positional encoding based on its original spatial coordinate, thereby preserving consistency between positional information and the image layout.

The backbone then operates on the compressed token sequence defined by $\mathcal{I}_{\text{keep}}$. When SD is enabled, NTP, MTP, and TCR are all adapted to this compressed sequence during training: specifically, the prediction targets of NTP and MTP remain defined with respect to the original sequence indices, while TCR samples token features from the retained hidden features.

\begin{algorithm}[t]
\caption{Pseudocode of the MTAR Training Framework}
\label{algo:overview}
\begin{algorithmic}[1]
\small
\STATE \textbf{Input:} image patches $\{x_i\}$; condition $c$;
        number of kept patches $M$;
        MTP offsets $\{\delta_w\}_{w=1}^{W}$;
        contrastive sample number $K$;
        TCR temperature $\tau$;
        loss weights $\lambda_{\mathrm{MTP}}, \lambda_{\mathrm{TCR}}$.
\STATE \textbf{Init:} Backbone; 2D RoPE positional encoding $\mathrm{Pos2D}$;
        MTP refinement heads $\{\mathrm{head}_{\mathrm{MTP}}^{w}\}_{w=1}^{W}$; MLP $f_\theta$.
\vspace{2pt}\hrule\vspace{2pt}
\STATE \textcolor{gray}{// \textit{Semantic Drop}}
\STATE $\mathcal{I}_{\mathrm{keep}} \leftarrow$ sorted indices of $M$ patches sampled from the temperature-scaled semantic importance distribution
\STATE $H \leftarrow \mathrm{Backbone}\!\left(c,\{x_i\}_{i\in\mathcal{I}_{\mathrm{keep}}},\mathrm{Pos2D}[\mathcal{I}_{\mathrm{keep}}]\right)$
\STATE Construct valid prediction masks from $\mathcal{I}_{\mathrm{keep}}$
\STATE \textcolor{gray}{// \textit{NTP Loss}}
\STATE Compute $L_{\mathrm{NTP}}$ on valid positions according to Eq.~(\ref{eq:ntp_loss})
\STATE \textcolor{gray}{// \textit{MTP Loss}}
\STATE $L_{\mathrm{MTP}} \leftarrow 0$
\FOR{$w=1$ \textbf{to} $W$}
    \STATE $H^{\mathrm{en}}_w \leftarrow \mathrm{head}_{\mathrm{MTP}}^{\,w}(H)$
    \STATE Update $L_{\mathrm{MTP}}$ on valid positions with the $w$-th term in Eq.~(\ref{eq:mtp_loss})
\ENDFOR
\STATE \textcolor{gray}{// \textit{TCR Loss}}
\STATE $H_{\mathrm{batch}} \leftarrow$ stack the hidden features from all samples in the batch
\STATE $\mathbf{F} \leftarrow \mathrm{RandomSample}\!\left(H_{\mathrm{batch}}[:,1:,:],\,K\right)$
\STATE Sample two independent dropout masks $m_1,m_2$
\STATE $\mathbf{Z}_1 \leftarrow \mathrm{L2Norm}(f_\theta(\mathrm{Dropout}(\mathbf{F};\,m_1)))$
\STATE $\mathbf{Z}_2 \leftarrow \mathrm{L2Norm}(f_\theta(\mathrm{Dropout}(\mathbf{F};\,m_2)))$
\STATE $\mathbf{S} \leftarrow \mathbf{Z}_1\mathbf{Z}_2^\top / \tau$
\STATE Compute $L_{\mathrm{TCR}}$ according to Eq.~(\ref{eq:tcr_loss})
\vspace{2pt}\hrule\vspace{2pt}
\STATE \textbf{Return:} $L_{\mathrm{NTP}}+\lambda_{\mathrm{MTP}}L_{\mathrm{MTP}}+\lambda_{\mathrm{TCR}}L_{\mathrm{TCR}}$
\end{algorithmic}
\end{algorithm}

\begin{table*}[t]
\caption{Model comparisons on the class-conditional ImageNet $256\times256$ benchmark~\cite{deng2009imagenet}. Metrics include Fr\'echet inception distance (FID), inception score (IS), precision, and recall. ``$\downarrow$'' and ``$\uparrow$'' indicate that lower and higher values are better, respectively.  $^{\dagger}$ denotes results reported by NAR~\cite{he2025neighboring}. ``-re'' denotes the use of rejection sampling.}

\label{tab:mainResult}

\centering
\normalsize
\setlength{\tabcolsep}{11pt}
\renewcommand{\arraystretch}{1.20}
\begin{tabular}{c|l c|c c c c}
\toprule
Type & Model & \#Para. & FID$\downarrow$ & IS$\uparrow$ & Precision$\uparrow$ & Recall$\uparrow$ \\
\midrule
\multirow{2}{*}{GAN}
& BigGAN~\cite{brock2018large}                & 112M & 6.95  & 224.5 & 0.89 & 0.38 \\
& GigaGAN~\cite{kang2023scaling}             & 569M & 3.45  & 225.5 & 0.84 & 0.61 \\
\midrule
\multirow{4}{*}{Diffusion}
& ADM~\cite{dhariwal2021diffusion}           & 554M & 10.94 & 101.0 & 0.69 & 0.63 \\
& CDM~\cite{ho2022cascaded}                  & --   & 4.88  & 158.7 & --   & --   \\
& LDM-4-G~\cite{ldm}               & 400M & 3.60  & 247.7 & 0.87 & 0.48 \\
& DiT-L/2~\cite{Peebles_2023_ICCV}           & 458M & 5.02  & 167.2 & 0.75 & 0.57 \\
\midrule
\multirow{3}{*}{Mask}
& MaskGIT~\cite{chang2022maskgit}            & 227M & 6.18  & 182.1 & 0.80 & 0.51 \\
& MaskGIT-re~\cite{chang2022maskgit}         & 227M & 4.02  & 355.6 & --   & --   \\
& MAGE~\cite{Li_2023_CVPR}                   & 439M & 7.04  & 123.5 & --   & --   \\
\midrule
\multirow{14}{*}{AR}
& VQGAN~\cite{vqgan}                         & 227M & 18.65 & 80.4  & 0.78 & 0.26 \\
& VQGAN~\cite{vqgan}                         & 1.4B & 15.78 & 74.3  & --   & --   \\
& VQGAN-re~\cite{vqgan}                      & 1.4B & 5.20  & 280.3 & --   & --   \\
& RQTran.~\cite{lee2022autoregressive}       & 3.8B & 7.55  & 134.0 & --   & --   \\
& RQTran.-re~\cite{lee2022autoregressive}    & 3.8B & 3.80  & 323.7 & --   & --   \\
& ViT-VQGAN~\cite{yu2022vector}              & 1.7B & 4.17  & 175.1 & --   & --   \\
& DART-AR~\cite{gu2025dart}                  & 812M & 3.98  & 256.8 & --   & --   \\
& LlamaGen-B~\cite{sun2024autoregressive}    & 111M & 5.46  & 193.61 & 0.84 & 0.46 \\
& LlamaGen-L~\cite{sun2024autoregressive}    & 343M & 3.80  & 248.28 & 0.83 & 0.52 \\
& IAR-B~\cite{Hu_2025_CVPR}                  & 111M & 5.14  & 202.0 & 0.85 & 0.45 \\
& IAR-L~\cite{Hu_2025_CVPR}                  & 343M & 3.18  & 234.8 & 0.82 & 0.53 \\
& PAR-L-4X$^{\dagger}$~\cite{Wang_2025_CVPR} & 343M & 4.32  & 189.4 & 0.87 & 0.43 \\
& NAR-B~\cite{he2025neighboring}                & 130M & 4.65  & 212.3 & 0.83 & 0.47 \\
& NAR-L~\cite{he2025neighboring}                & 372M & 3.06  & 263.9 & 0.81 & 0.53 \\
\midrule
\multirow{2}{*}{Our}
& MTAR-B                                     & 138M & 4.50  & 213.96 & 0.84 & 0.48 \\
& MTAR-L                                     & 387M & 2.85  & 271.41 & 0.83 & 0.54 \\
\bottomrule
\end{tabular}
\end{table*}

\section{Experiments}

\subsection{Experimental Settings}

\noindent\textbf{Implementation Details.}
Based on the MTAR framework constructed upon LlamaGen~\cite{sun2024autoregressive}, we adopt LlamaGen~\cite{sun2024autoregressive} as our primary baseline and systematically evaluate its performance on the class-conditional image generation task under the standard ImageNet $256\times256$ generation benchmark~\cite{deng2009imagenet}. We employ the VQGAN~\cite{vqgan}  tokenizer pre-trained on ImageNet from LlamaGen~\cite{sun2024autoregressive}, with a vocabulary size of $16{,}384$.  For $256\times256$ resolution images, the tokenizer performs a $16\times$ downsampling. We use the AdamW~\cite{kingma2014adam,luo2024open} optimizer with the following hyperparameters:  
$\beta_1 = 0.9$, $\beta_2 = 0.95$, weight decay $= 0.05$, and gradient clipping threshold $= 1.0$.  
Dropout rates for the input embedding layer, attention modules, and feed-forward networks (FFN) are all set to $0.1$.  During inference, we follow the traditional autoregressive generation 
procedure of LlamaGen~\cite{sun2024autoregressive} without any modification. For SD, the semantic importance scores are extracted offline once before training and reused across all epochs, incurring only a one-time preprocessing overhead.

\noindent\textbf{Dataset and Evaluation Protocol.}
For ablation studies, we construct a small-scale training subset by sampling 100k images from ImageNet, denoted as ImageNet-100k $256\times256$~\cite{deng2009imagenet}, for efficient training. We then generate 10k images for evaluation against 10k real images sampled from the ImageNet validation set. For full-scale evaluation, we train on the complete ImageNet dataset and generate 50k images for assessment, following the same validation protocol as LlamaGen~\cite{sun2024autoregressive}. We report FID~\cite{heusel2017gans}, IS~\cite{salimans2016improved}, Precision, and Recall~\cite{kynkaanniemi2019improved}, all computed using the ADM TensorFlow evaluation suite~\cite{dhariwal2021diffusion}.

\begin{table*}[t]
\caption{Comparison between LlamaGen~\cite{sun2024autoregressive} and MTAR in terms of generation quality and training efficiency. Speedup is measured with respect to the total training time of the baseline LlamaGen~\cite{sun2024autoregressive}.}
\label{tab:speedup}
\centering
\normalsize
\setlength{\tabcolsep}{10pt}
\renewcommand{\arraystretch}{1.20}
\begin{tabular}{l|c|ccccc}
\toprule
Model & Epochs & FID$\downarrow$ & IS$\uparrow$ & Precision$\uparrow$ & Recall$\uparrow$ & Speedup \\
\midrule
LlamaGen-B~\cite{sun2024autoregressive} & 300 & 5.46 & 193.61 & 0.84 & 0.46 & 1.00$\times$ \\
\midrule
\multirow{4}{*}{MTAR-B}
& 50  & 5.76 & 198.70 & 0.86 & 0.42 & 7.62$\times$ \\
& 100 & 5.37 & 209.07 & 0.86 & 0.44 & 3.81$\times$ \\
& 250 & 4.66 & 211.54 & 0.84 & 0.47 & 1.73$\times$ \\
& 300 & 4.50 & 213.96 & 0.84 & 0.48 & 1.27$\times$ \\
\midrule
LlamaGen-L~\cite{sun2024autoregressive} & 300 & 3.80 & 248.28 & 0.83 & 0.52 & 1.00$\times$ \\
\midrule
\multirow{4}{*}{MTAR-L}
& 50  & 3.74 & 219.29 & 0.82 & 0.52 & 8.33$\times$ \\
& 100 & 3.36 & 226.30 & 0.81 & 0.54 & 4.17$\times$ \\
& 250 & 2.96 & 267.46 & 0.83 & 0.55 & 1.88$\times$ \\
& 300 & 2.85 & 271.41 & 0.83 & 0.54 & 1.39$\times$ \\
\bottomrule
\end{tabular}
\end{table*}

\begin{table*}[t]

\caption{Ablation of MTP head directions based on the NTP head (R$_1$). We compare auxiliary prediction along three directions: right, bottom, and bottom-right.}
\label{tab:ablation MTP direction}
\centering
\normalsize
\setlength{\tabcolsep}{5pt}
\renewcommand{\arraystretch}{1.2}
\begin{tabular}{l|ccc|c|cccc}
\toprule
Method & Right (R$_2$) & Bottom (B) & Bottom-Right (B.R$_1$) & \#Para. & FID$\downarrow$ & IS$\uparrow$ & Precision$\uparrow$ & Recall$\uparrow$ \\
\midrule
MTP (R$_1$, R$_2$)   & \checkmark &            &            & 138M & 20.13 & 59.32 & 0.75 & 0.45 \\
MTP (R$_1$, B)       &            & \checkmark &            & 138M & 19.75 & 62.12 & 0.75 & 0.45 \\
MTP (R$_1$, B.R$_1$) &            &            & \checkmark & 138M & 19.96 & 61.05 & 0.75 & 0.44 \\
\bottomrule
\end{tabular}
\end{table*}



\begin{table}[t]

\centering

\caption{ Ablation study of MTP, TCR, and SD on LlamaGen-B~\cite{sun2024autoregressive}.}

\label{tab:ablation MTP TCR SD}
\setlength{\tabcolsep}{3pt}
\renewcommand{\arraystretch}{1.2}
\begin{tabular}{l|c|ccccc}
\toprule
Method & \#Para. & FID$\downarrow$ & IS$\uparrow$ & Precision$\uparrow$ & Recall$\uparrow$ & Speedup \\
\midrule
LlamaGen-B & 111M & 21.96 & 54.04 & 0.71 & 0.46 & 1.00$\times$ \\
\midrule
MTP & 138M & 19.75 & 62.12 & 0.75 & 0.45 & 0.79$\times$ \\
TCR & 111M & 20.55 & 57.78 & 0.72 & 0.47 & 0.98$\times$ \\
SD  & 111M & 22.02 & 54.99 & 0.70 & 0.47 & 1.62$\times$ \\
\midrule
MTP+TCR    & 138M & 18.78 & 64.33 & 0.73 & 0.48 & 0.79$\times$ \\
MTP+SD     & 138M & 20.79 & 66.17 & 0.71 & 0.47 & 1.27$\times$ \\
TCR+SD     & 111M & 20.14 & 69.07 & 0.73 & 0.46 & 1.61$\times$ \\
MTP+TCR+SD & 138M & 18.65 & 66.32 & 0.75 & 0.46 & 1.27$\times$ \\
\bottomrule
\end{tabular}
\end{table}

\begin{table}[t]
\caption{ Ablation on the number of MTP heads. Performance does not consistently improve as more auxiliary heads are added.}
\label{tab:ablation MTP num}
\centering
\normalsize
\setlength{\tabcolsep}{3pt}
\renewcommand{\arraystretch}{1.2}
\begin{tabular}{l|c|cccc}
\toprule
Method & \#Para. & FID$\downarrow$ & IS$\uparrow$ & Precision$\uparrow$ & Recall$\uparrow$ \\
\midrule
MTP\ (R$_1$, R$_2$, R$_3$)              & 164M & 19.79 & 41.77 & 0.74 & 0.47 \\
MTP\ (R$_1$, R$_2$, B)                  & 164M & 20.34 & 59.85 & 0.72 & 0.46 \\
MTP\ (R$_1$, B, B.R$_1$)                & 164M & 19.94 & 62.33 & 0.74 & 0.46 \\
\midrule
MTP\ (R$_1$, R$_2$, R$_3$, R$_4$)       & 191M & 20.95 & 57.15 & 0.73 & 0.45 \\
MTP\ (R$_1$, R$_2$, B, B.R$_1$)         & 191M & 20.95 & 57.18 & 0.73 & 0.45 \\
\midrule
MTP\ (R$_1$, R$_2$, R$_3$, R$_4$, R$_5$)& 218M & 20.18 & 59.46 & 0.74 & 0.45 \\
\bottomrule
\end{tabular}
\end{table}

\begin{table}[t]

\caption{Ablation study on the number of sampled tokens for TCR. We observe that using 2,048 sampled tokens yields the best overall performance.}
\label{tab:ablation TCR tokens num}

\centering
\normalsize
\renewcommand{\arraystretch}{1.2}
\begin{tabular*}{\columnwidth}{@{\extracolsep{\fill}}c|cccc}
\toprule
Sampled Tokens & FID$\downarrow$ & IS$\uparrow$ & Precision$\uparrow$ & Recall$\uparrow$ \\
\midrule
16384 & 20.70 & 58.60 & 0.73 & 0.46 \\
8192  & 19.83 & 58.62 & 0.73 & 0.47 \\
4096  & 19.71 & 57.82 & 0.72 & 0.48 \\
2048  & 18.78 & 64.33 & 0.73 & 0.48 \\
1024  & 19.18 & 63.54 & 0.73 & 0.48 \\
512   & 18.94 & 64.45 & 0.74 & 0.48 \\
256   & 19.04 & 64.76 & 0.74 & 0.47 \\
\bottomrule
\end{tabular*}
\end{table}

\begin{table}[t]

\caption{Ablation on drop strategies. DINOv3~\cite{simeoni2025dinov3}-guided SD achieves the best performance.}
\label{tab:ablation Drop Criterion}

\centering
\normalsize
\renewcommand{\arraystretch}{1.2}
\begin{tabular*}{\columnwidth}{@{\extracolsep{\fill}}c|cccc}
\toprule
Drop Strategy & FID$\downarrow$ & IS$\uparrow$ & Precision$\uparrow$ & Recall$\uparrow$ \\
\midrule
Random & 19.97 & 62.69 & 0.72 & 0.45 \\
SigLIP2~\cite{tschannen2025siglip} & 19.70 & 63.84 & 0.73 & 0.46 \\
DINOv3~\cite{simeoni2025dinov3} & 18.65 & 66.32 & 0.75 & 0.46 \\
\bottomrule
\end{tabular*}
\end{table}

\begin{table}[t]

\caption{Ablation on the patch drop rate in SD. A 50\% drop rate achieves the best overall trade-off. Speedup is measured relative to the MTP+TCR configuration without SD.}
\label{tab:ablation drop rate}

\centering
\normalsize
\renewcommand{\arraystretch}{1.2}
\begin{tabular*}{\columnwidth}{@{\extracolsep{\fill}}c|ccccc}
\toprule
Drop Rate & FID$\downarrow$ & IS$\uparrow$ & Precision$\uparrow$ & Recall$\uparrow$ & Speedup \\
\midrule
60\% & 19.31 & 62.93 & 0.74 & 0.45 & 1.81$\times$ \\
50\% & 18.65 & 66.32 & 0.75 & 0.46 & 1.60$\times$ \\
40\% & 18.91 & 65.32 & 0.74 & 0.46 & 1.38$\times$ \\
\bottomrule
\end{tabular*}
\end{table}

\begin{table}[t]

\caption{Ablation on the two-stage SD training schedule. The 80\%:20\% schedule provides a favorable quality--efficiency trade-off. Speedup is measured relative to the MTP+TCR configuration without SD.}
\label{tab:ablation SD Schedule}

\centering
\normalsize
\renewcommand{\arraystretch}{1.2}
\begin{tabular*}{\columnwidth}{@{\extracolsep{\fill}}c|ccccc}
\toprule
SD Schedule & FID$\downarrow$ & IS$\uparrow$ & Precision$\uparrow$ & Recall$\uparrow$ & Speedup \\
\midrule
40\%:60\% & 18.65 & 65.54 & 0.75 & 0.46 & 1.23$\times$ \\
50\%:50\% & 18.60 & 65.01 & 0.74 & 0.48 & 1.31$\times$ \\
60\%:40\% & 18.41 & 64.85 & 0.74 & 0.48 & 1.39$\times$ \\
70\%:30\% & 18.68 & 64.38 & 0.74 & 0.48 & 1.49$\times$ \\
80\%:20\% & 18.65 & 66.32 & 0.75 & 0.46 & 1.60$\times$ \\
90\%:10\% & 19.40 & 62.89 & 0.75 & 0.46 & 1.73$\times$ \\
\bottomrule
\end{tabular*}
\end{table}

\subsection{Main Results}

We conduct systematic comparisons between MTAR and a diverse set of representative generative models on the ImageNet $256\times256$ benchmark~\cite{deng2009imagenet}, including GAN-based methods~\cite{brock2018large,kang2023scaling}, diffusion-based methods~\cite{dhariwal2021diffusion,ho2022cascaded,ding2023patched,Peebles_2023_ICCV}, masked prediction methods~\cite{chang2022maskgit,Li_2023_CVPR}, and autoregressive methods~\cite{vqgan,lee2022autoregressive,yu2022vector,gu2025dart,sun2024autoregressive,Hu_2025_CVPR,Wang_2025_CVPR}. As shown in Tab~\ref{tab:mainResult}, MTAR is highly competitive among autoregressive approaches. Specifically, MTAR-B achieves an FID of 4.50 with only 138M parameters, outperforming multiple autoregressive baselines at a comparable scale. MTAR-L further improves the FID to 2.85, achieving the best result among autoregressive models of similar size.

\begin{figure}[t]
\centering
\includegraphics[width=1\linewidth]{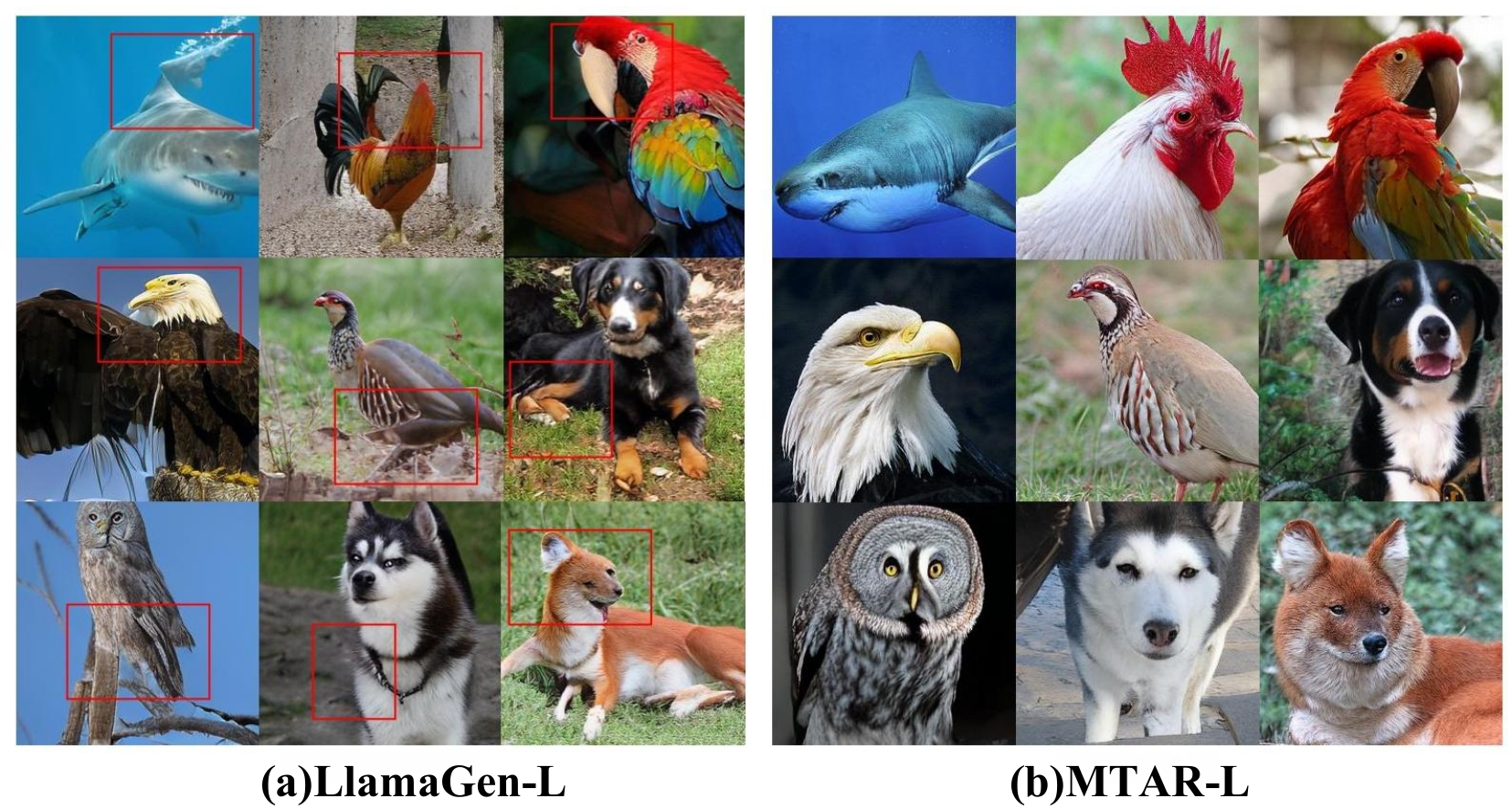}
\caption{Qualitative comparison between LlamaGen-L~\cite{sun2024autoregressive} and MTAR-L on ImageNet $256\times256$~\cite{deng2009imagenet}. Under the same class conditions, MTAR-L produces images with clearer semantics, more coherent structures, and fewer artifacts. Red boxes highlight representative artifacts in LlamaGen-L~\cite{sun2024autoregressive} samples.}
\Description{ImageNet samples generated by LlamaGen-L and MTAR-L, with red
boxes marking artifacts in the LlamaGen-L results.}
\label{fig:visual}
\end{figure}

Compared with LlamaGen~\cite{sun2024autoregressive}, MTAR consistently achieves better generation quality and training efficiency, as shown in Tab~\ref{tab:speedup}. Under the same number of training iterations, MTAR-B and MTAR-L improve FID by 0.96 and 0.95, respectively, while achieving training speedups of 1.27$\times$ and 1.39$\times$. Even under smaller training budgets, MTAR remains superior: MTAR-B already surpasses LlamaGen-B~\cite{sun2024autoregressive} with only about one-third of the training iterations, while MTAR-L outperforms its corresponding baseline with only about one-sixth of the training iterations, reaching speedups of 3.81$\times$ and 8.33$\times$, respectively. As shown in Fig~\ref{fig:visual}, MTAR-L generates samples with more coherent local structures and fewer repetitive texture artifacts than LlamaGen-L~\cite{sun2024autoregressive}, suggesting reduced degeneration such as texture repetition and local collapse under the same class conditions.

\begin{figure}[t]
\centering
\includegraphics[width=1\linewidth]{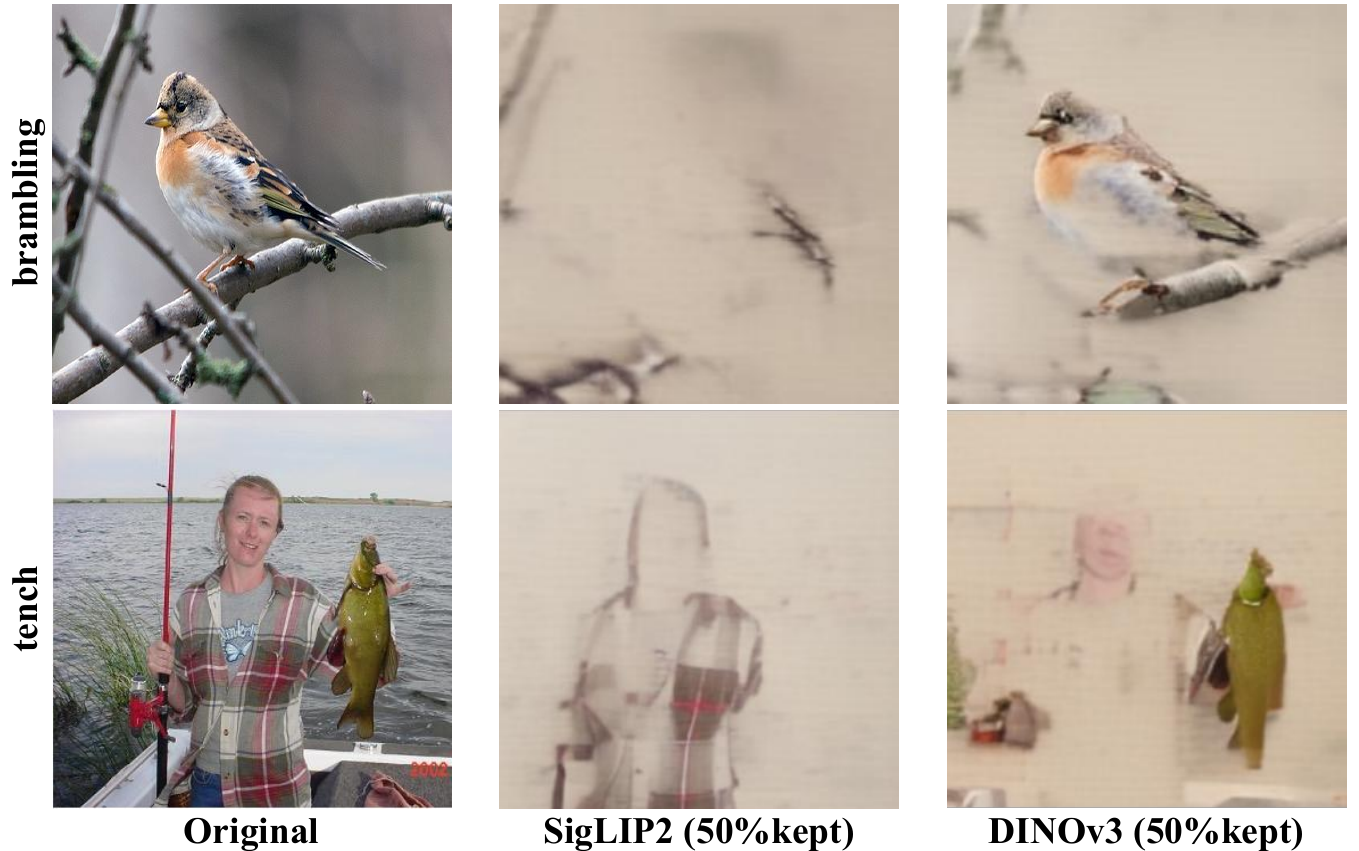}

\caption{We retain the top-50\% patches according to the importance scores produced by SigLIP2~\cite{tschannen2025siglip} and DINOv3~\cite{simeoni2025dinov3}, and visualize the corresponding VQ-decoded reconstructions. Compared with SigLIP2~\cite{tschannen2025siglip}, DINOv3~\cite{simeoni2025dinov3} better preserves the semantic content and structural integrity of the foreground object.}
\Description{VQ-decoded reconstructions using the top-50\% patches selected
by SigLIP2 and DINOv3, with DINOv3 preserving foreground objects more clearly.}
\label{fig:Semantic Drop Dinov3 siglip2}

\end{figure}

\subsection{Ablation Studies}
We conduct ablation studies to examine (i) whether MTP, TCR, and SD address
the corresponding limitations of traditional NTP, (ii) which design choices
are critical to each component, and (iii) how MTAR balances generation quality
and training efficiency. Unless otherwise specified, all experiments use
MTAR-B trained for 50 epochs on ImageNet-100k at
$256\times256$~\cite{deng2009imagenet}. In the SD ablations, all factors
except the one under study remain at their default values.
Tabs~\ref{tab:ablation Drop Criterion} and~\ref{tab:ablation drop rate}
use the two-stage schedule SD (80\%:20\%), which applies a 50\% patch drop
rate during the first 80\% of training and full-sequence training during the
remaining 20\%. We evaluate component combinations; the prediction direction
and auxiliary-head count of MTP; the sampled-token count of TCR; and the
importance encoder, patch drop rate, and training schedule of SD.

\noindent\textbf{Ablation of MTP, TCR, and SD.}
Tab~\ref{tab:ablation MTP TCR SD} evaluates the individual and combined effects of MTP, TCR, and SD.
Starting from LlamaGen-B~\cite{sun2024autoregressive}, MTP yields the largest individual improvement, reducing the FID from 21.96 to 19.75 and increasing the IS from 54.04 to 62.12, at the cost of a lower speedup of 0.79$\times$.
TCR also improves the baseline, achieving an FID of 20.55 while largely preserving the baseline speed (0.98$\times$).
Although SD alone does not improve FID (22.02), it substantially improves efficiency, achieving a speedup of 1.62$\times$.
When combined with MTP, TCR further reduces the FID to 18.78 and improves Recall to 0.48, without increasing the parameter count beyond that of MTP.
Finally, adding SD to MTP and TCR yields the best FID of 18.65, while retaining a 1.27$\times$ speedup over the baseline.
These results show that MTP provides the primary quality gain, TCR offers complementary regularization, and SD effectively improves efficiency in the full configuration.

\noindent\textbf{Prediction direction and number of heads in MTP.}
We examine the effects of auxiliary prediction directions and head counts.
As shown in Tabs~\ref{tab:ablation MTP direction}
and~\ref{tab:ablation MTP num}, with one auxiliary head, MTP (R$_1$, B)
achieves the best FID of 19.75, compared with 20.13 for MTP
(R$_1$, R$_2$) and 19.96 for MTP (R$_1$, B.R$_1$). Adding more heads
provides no consistent improvement: MTP (R$_1$, R$_2$, R$_3$) and MTP
(R$_1$, B, B.R$_1$) obtain FIDs of 19.79 and 19.94, respectively, while
MTP (R$_1$, R$_2$, R$_3$, R$_4$) further degrades to 20.95.

These results show that MTP benefits from effective auxiliary supervision
rather than simply predicting more future tokens. Unlike language modeling,
where MTP extends prediction along a 1D sequence, image generation benefits
from supervision aligned with its 2D structure. Since NTP already captures
horizontal dependencies, predicting the spatially lower neighbor provides
less redundant information. Together with 2D positional encoding, this
direction helps the shared representations capture richer spatial patterns
and future context beyond raster order. We therefore adopt MTP
(R$_1$, B) as the default configuration.

\noindent\textbf{Effect of the number of sampled tokens in TCR.}
Tab~\ref{tab:ablation TCR tokens num} shows that 2,048 sampled tokens achieve the best performance, with an FID of 18.78. Smaller sampling sizes (256, 512, and 1,024) yield comparable but slightly worse results, whereas larger sizes (8,192 and 16,384) degrade performance, increasing FID to 19.83 and 20.70, respectively. This suggests that small sampling sizes provide insufficient contrastive coverage, while excessively large sizes introduce redundant or highly correlated token pairs that weaken regularization. We therefore use 2,048 sampled tokens by default.

\noindent\textbf{Design choices of Semantic Dropping.}
We study three key factors of SD: the semantic importance estimator, patch
drop rate, and two-stage training schedule. As shown in
Tab~\ref{tab:ablation Drop Criterion}, semantic-aware dropping consistently
outperforms random dropping, with DINOv3~\cite{simeoni2025dinov3} reducing
FID from 19.97 to 18.65. Fig~\ref{fig:Semantic Drop Dinov3 siglip2} further
shows that DINOv3 preserves foreground objects and their structures better
than SigLIP2~\cite{tschannen2025siglip}. According to
Tab~\ref{tab:ablation drop rate}, a 50\% drop rate achieves the best
quality--efficiency trade-off, with an FID of 18.65 and a
1.60$\times$ speedup. A 40\% drop rate provides less acceleration, whereas
a 60\% rate is faster but noticeably degrades generation quality. We
therefore use DINOv3 with a 50\% drop rate by default.

As shown in Tab~\ref{tab:ablation SD Schedule}, SD (60\%:40\%) obtains the
lowest FID of 18.41, while SD (80\%:20\%) offers a better balance, achieving
an FID of 18.65 with a 1.60$\times$ speedup. A shorter recovery stage, such
as SD (90\%:10\%), increases speed but reduces generation quality. These
results suggest applying SD during the early and middle training stages
while retaining a sufficiently long final full-sequence stage to recover
fine-grained token dependencies and stabilize performance. We therefore
adopt SD (80\%:20\%) as the default schedule.

\section{Conclusion}
In this paper, we present MTAR, a unified training framework for autoregressive image generation. MTAR improves conventional next-token prediction through denser supervision, more discriminative token representations, and more efficient computation, via MTP, TCR, and SD, respectively. All three components are training-only and leave the autoregressive inference pipeline unchanged.
Experiments on ImageNet $256\times256$ show that MTAR achieves a better trade-off between generation quality and training efficiency. Compared with LlamaGen, MTAR achieves better FID with substantially faster training, and remains effective under smaller training budgets. Ablation studies further verify the effectiveness of each component and its key design choices.
Overall, MTAR demonstrates that autoregressive image generation can be improved through better training design, without modifying the inference pipeline.

\bibliographystyle{ACM-Reference-Format}
\bibliography{sample-base}





\title[MTAR Supplementary Material]{
Efficient Training with Foresight: Multi-Token Auxiliary
Supervision for Autoregressive Image Generation
}

\appendix

\twocolumn[
\begin{center}
    {\LARGE\bfseries
    Efficient Training with Foresight: Multi-Token Auxiliary\\
    Supervision for Autoregressive Image Generation\par}
    \medskip
    {\LARGE Supplementary Material\par}
    \bigskip
\end{center}
]

\section*{Overview}
The supplementary material is organized as follows:
\begin{itemize}
    \item Sec.~\ref{sec:A} provides implementation and hyper-parameter details.
    \item Sec.~\ref{sec:B} presents additional ablations on SD schedules,
    transfer to RAR, MTP-head optimization, and CFG sensitivity.
    \item Sec.~\ref{sec:C} analyzes token discriminability, potential false
    negatives, and the qualitative behavior of Semantic Dropping.
    \item Sec.~\ref{sec:D} provides additional samples generated by MTAR.
    \item Sec.~\ref{sec:E} compares samples generated by MTAR and
    LlamaGen.
\end{itemize}

\section{Implementation Details}
\label{sec:A}

The training and sampling hyper-parameters used for MTAR are summarized in
Tab~\ref{tab:mtar_hparams}.

\begin{table}[h]
\centering
\caption{Detailed training and sampling hyper-parameters of MTAR.}
\label{tab:mtar_hparams}
\begin{tabular}{p{0.45\columnwidth}|p{0.45\columnwidth}}
config & value \\
\Xhline{1.2pt}
\multicolumn{2}{c}{\textit{training hyper-parameters}} \\
\Xhline{0.8pt}
optimizer & AdamW \\
learning rate & 1e-4 \\
weight decay & 5e-2 \\
optimizer momentum & $(0.9, 0.95)$ \\
batch size & 256 \\
total epochs & 300 \\
precision & bfloat16 \\
max grad norm & 1.0 \\
dropout rate & 0.1 \\
downsample size & 16 \\
$\lambda_{MTP}$ & 0.1 \\
$\lambda_{TCR}$ & 0.2 \\
TCR temperature & 0.07 \\
TCR dropout rate & 0.2 \\
TCR sample tokens & 2048 \\
semantic patch drop rate & 0.50 \\
semantic drop temperature & 0.35 \\
semantic drop schedule & two-stage (80\% / 20\%) \\
\Xhline{0.8pt}
\multicolumn{2}{c}{\textit{sampling hyper-parameters}} \\
\Xhline{0.8pt}
temperature & 1.0 \\
guidance scale & 2.00 (B, L) \\
\Xhline{0.8pt}
\end{tabular}
\end{table}

\section{Additional Ablation Studies}
\label{sec:B}

This section supplements the main-paper ablations with details on the
evaluation protocol, three-stage SD schedules, transfer to another
autoregressive baseline, optimization with multiple MTP heads, and CFG
sensitivity.

\subsection{Ablation Protocol}

We use ImageNet-100K and 50-epoch training for data-efficient ablations that
isolate individual design choices. We additionally validate MTAR under the
full ImageNet 300-epoch setting, confirming that the conclusions transfer
across data and training scales.

\subsection{Three-Stage SD Schedules}

Table~\ref{tab:ablation_SD_three_stage} compares several three-stage SD
schedules. The notation $a\%{:}b\%{:}c\%$ denotes the relative durations of
the three stages, which use patch-drop rates of 50\%, 25\%, and 0\%,
respectively. Although the
60\%:5\%:35\% schedule achieves the best FID among the three-stage variants,
Tab.~\ref{tab:ablation SD Schedule} shows that the simpler two-stage
80\%:20\% schedule provides a better overall trade-off between generation
quality and training efficiency. We therefore adopt the two-stage schedule
in the final model.

\begin{table}[h]
\caption{Ablation study on three-stage SD schedules. The three entries denote
the relative durations of stages with patch-drop rates of 50\%, 25\%, and
0\%, respectively. Different schedules lead to different quality--efficiency
trade-offs, and 60\%:5\%:35\% achieves the best FID among the evaluated
three-stage variants.}
\label{tab:ablation_SD_three_stage}
\centering
\normalsize
\renewcommand{\arraystretch}{1.2}
\begin{tabular*}{\columnwidth}{@{\extracolsep{\fill}}l|ccccc}
\toprule
SD Schedule & FID$\downarrow$ & IS$\uparrow$ & Precision$\uparrow$ &
Recall$\uparrow$ & Speedup$\uparrow$ \\
\midrule
40\%:5\%:55\%  & 18.62 & 66.47 & 0.74 & 0.47 & 1.25$\times$ \\
40\%:10\%:50\% & 18.77 & 63.01 & 0.75 & 0.46 & 1.27$\times$ \\
50\%:5\%:45\%  & 18.56 & 67.11 & 0.75 & 0.46 & 1.33$\times$ \\
50\%:10\%:40\% & 18.99 & 65.04 & 0.75 & 0.46 & 1.35$\times$ \\
60\%:5\%:35\%  & 18.20 & 68.25 & 0.75 & 0.47 & 1.42$\times$ \\
60\%:10\%:30\% & 18.56 & 65.77 & 0.74 & 0.46 & 1.44$\times$ \\
70\%:5\%:25\%  & 18.61 & 66.60 & 0.76 & 0.46 & 1.52$\times$ \\
70\%:10\%:20\% & 18.81 & 66.03 & 0.75 & 0.46 & 1.55$\times$ \\
\bottomrule
\end{tabular*}
\end{table}

\subsection{Generalization to RAR}

To evaluate whether MTP generalizes beyond the LlamaGen backbone, we
incorporate MTP(B) into RAR-B~\cite{Yu_2025_ICCV}. We use the official RAR
implementation and replace the MaskGIT-VQGAN tokenizer with the LlamaGen
VQGAN for a consistent comparison.

\begin{table}[h]
\caption{Generalization of MTP(B) to RAR-B~\cite{Yu_2025_ICCV}.
Adding MTP(B) consistently improves FID, IS, precision, and recall,
demonstrating that multi-token auxiliary supervision transfers beyond the
LlamaGen backbone.}
\label{tab:rar_mtp_ablation}
\centering
\small
\setlength{\tabcolsep}{3.2pt}
\renewcommand{\arraystretch}{1.12}
\begin{tabular}{l|c|cccc}
\toprule
Method & \#Para. & FID$\downarrow$ & IS$\uparrow$ &
Precision$\uparrow$ & Recall$\uparrow$ \\
\midrule
RAR-B~\cite{Yu_2025_ICCV} & 285M & 18.70 & 72.28 & 0.71 & 0.50 \\
RAR-B+MTP(B) & 320M & 17.82 & 76.49 & 0.73 & 0.51 \\
\bottomrule
\end{tabular}
\end{table}

As shown in Tab~\ref{tab:rar_mtp_ablation}, MTP(B) improves RAR-B from
18.70 to 17.82 FID, demonstrating that the auxiliary prediction objective
generalizes beyond the primary LlamaGen baseline. We do not directly compare
with STAR because its code is unavailable, while IAR-2 introduces a tokenizer
retraining stage that would make a controlled comparison costly and
potentially unfair.

\subsection{MTP-Head Optimization}

Figure~\ref{fig:mtp_loss} compares the optimization behavior of different
MTP-head configurations. Among the evaluated prediction sets, MTP
$(R_1,B)$ achieves the lowest primary autoregressive loss $L(R_1)$,
indicating that supervision from the spatially lower token is particularly
effective. Adding more auxiliary heads does not provide a consistent further
improvement, suggesting that spatial alignment is more important than simply
increasing the number of prediction targets.

\begingroup
\setlength{\intextsep}{8pt}
\begin{figure}[H]
\centering
\includegraphics[width=\linewidth]{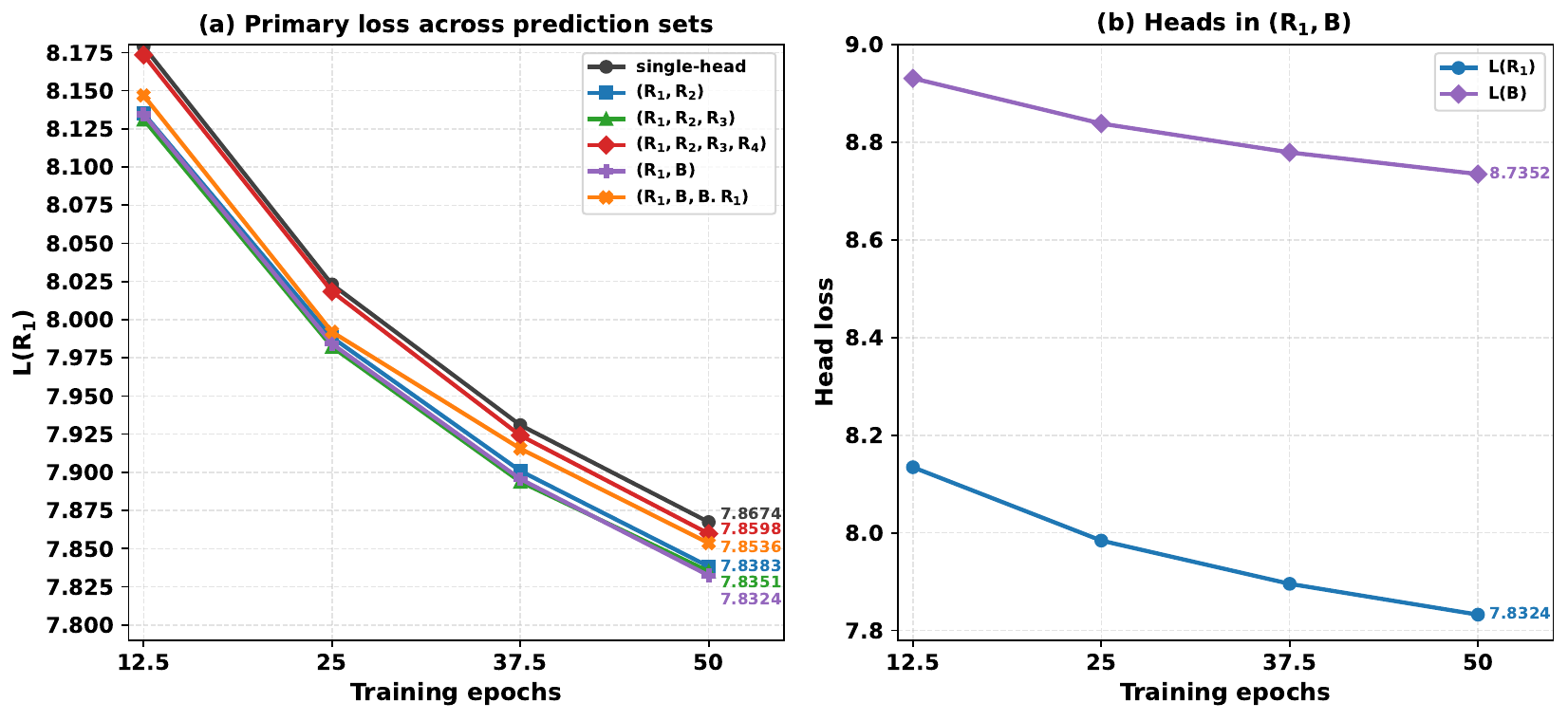}
\caption{Optimization behavior under different MTP-head configurations.
\textbf{(a)} Primary autoregressive loss $L(R_1)$ across training epochs for
single-head and multi-head prediction sets. \textbf{(b)} Primary-head loss
$L(R_1)$ and auxiliary-head loss $L(B)$ for the $(R_1,B)$ configuration.
The comparison illustrates how additional prediction targets alter the
optimization of the main autoregressive objective and motivates limiting the
number of auxiliary heads to avoid excessive interference.}
\label{fig:mtp_loss}
\end{figure}
\endgroup

\subsection{CFG Sensitivity}

We conduct a CFG sweep using the 300-epoch MTAR-B model. The resulting FID
values are 4.94, 4.50, and 4.85 at guidance scales of 1.75, 2.00, and 2.25,
respectively. We therefore use a guidance scale of 2.00.

\section{Representation and Semantic-Dropping Analysis}
\label{sec:C}

\subsection{Token Discriminability}

To evaluate whether TCR improves the discriminability of visual-token
representations, we conduct token-level linear probing and kNN-purity
analysis on the ImageNet validation set.

\begin{figure}[H]
\centering
\includegraphics[width=\linewidth]{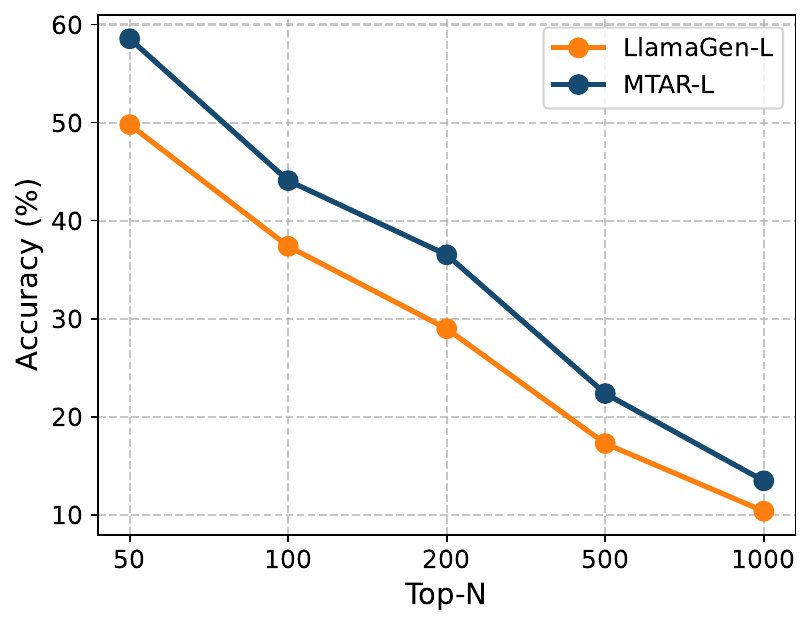}
\caption{Token-level linear-probing accuracy on the ImageNet validation set.
MTAR-L consistently outperforms LlamaGen-L
across all Top-$N$ settings, indicating better linear separability of
visual-token representations.}
\label{fig:linear_probing}
\end{figure}

For linear probing, we train a linear classifier to predict VQ token IDs from
frozen hidden representations under vocabulary sizes ranging from
Top-$N=50$ to Top-$N=1000$. As shown in Fig~\ref{fig:linear_probing}, MTAR-L
with TCR consistently outperforms LlamaGen-L without TCR, with gains ranging
from 3.10\% at Top-1000 to 8.76\% at Top-50. The larger improvements at
smaller $N$ suggest that TCR particularly enhances the representations of
frequent visual tokens.

For local neighborhood analysis, we compute token-level kNN purity with
$k=5$. For each token representation, we retrieve its five nearest neighbors
and measure the proportion sharing the same VQ token ID. MTAR-L achieves
17.54\% kNN purity, compared with 15.78\% for LlamaGen-L, an absolute
improvement of 1.76\%. These results indicate that TCR improves both global
linear separability and local neighborhood consistency.

\subsection{Potential False Negatives}

We use DINOv3~\cite{simeoni2025dinov3} to extract visual-token features from
10,000 ImageNet training images and compute pairwise cosine similarities.
As reported in Tab~\ref{tab:similarity}, only 2.01\% and 8.73\% of token
pairs have cosine similarity above 0.9 and 0.8, respectively. Highly similar
within-batch token pairs are therefore uncommon. Randomly sampling $K$ tokens
for contrastive learning and discarding 50\% of patch tokens through
Semantic Dropping further reduce the probability of false negatives.

\begin{table}[h]
\caption{Proportions of highly similar token pairs under different cosine
similarity thresholds.}
\label{tab:similarity}
\centering
\normalsize
\renewcommand{\arraystretch}{1.2}
\begin{tabular}{p{2.8cm}|p{1.8cm}}
\toprule
\centering Similarity \centering\arraybackslash &
\centering Ratio\arraybackslash \\
\midrule
\centering Sim $>$ 0.9 & \centering 2.01\%\arraybackslash \\
\centering Sim $>$ 0.8 & \centering 8.73\%\arraybackslash \\
\bottomrule
\end{tabular}
\end{table}

\subsection{Qualitative Behavior of Semantic Dropping}

\begin{figure}[h]
\centering
\includegraphics[width=0.95\linewidth]{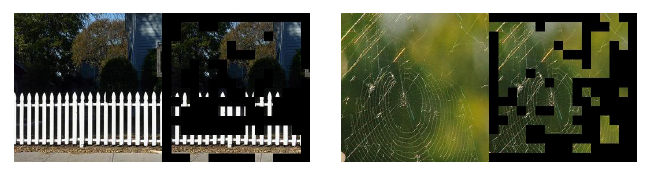}
\caption{Randomly selected examples before and after SD for the ImageNet
classes \textit{picket fence} and \textit{spider web}, using 50\% token
retention.}
\label{fig:sd_texture_cases}
\end{figure}

Figure~\ref{fig:sd_texture_cases} visualizes representative texture-rich
ImageNet samples before and after Semantic Dropping. With 50\% token
retention, SD removes substantial background redundancy while preserving
the primary foreground structures used for training.

\section{Additional Generated Samples}
\label{sec:D}

\begin{figure}[H]
\centering
\includegraphics[width=0.95\linewidth]{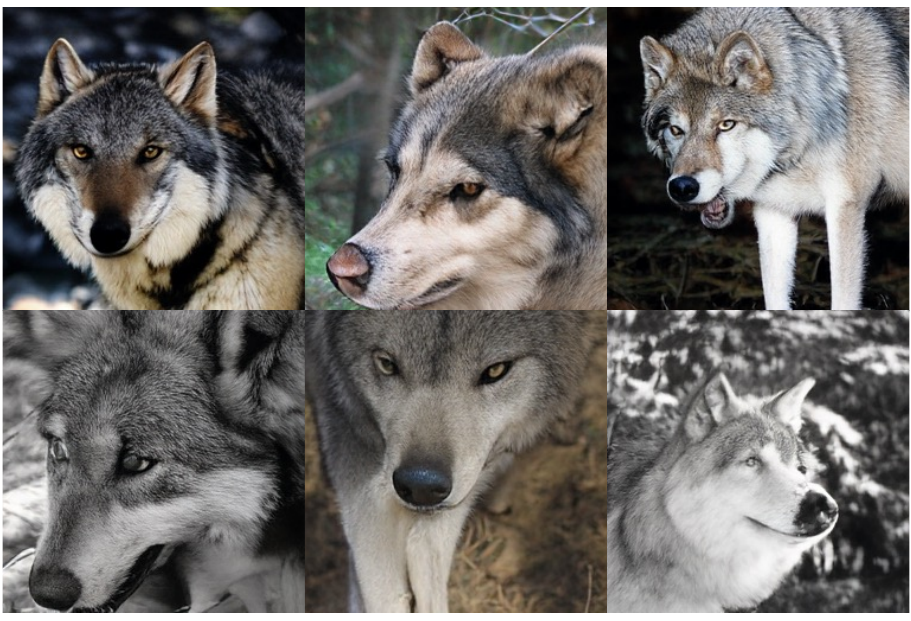}
\caption{Generated samples from MTAR-L for ImageNet class ID 269 (wolf).}
\label{fig:mtar_wolf}
\end{figure}

\begin{figure}[H]
\centering
\includegraphics[width=0.95\linewidth]{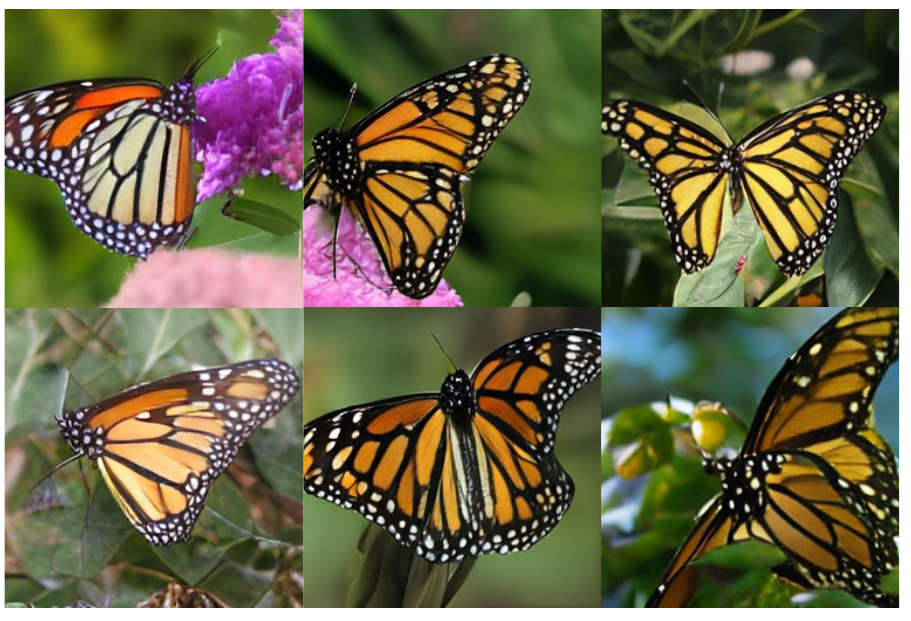}
\caption{Generated samples from MTAR-L for ImageNet class ID 323
(butterfly).}
\label{fig:mtar_butterfly}
\end{figure}

\begin{figure}[h]
\centering
\includegraphics[width=0.95\linewidth]{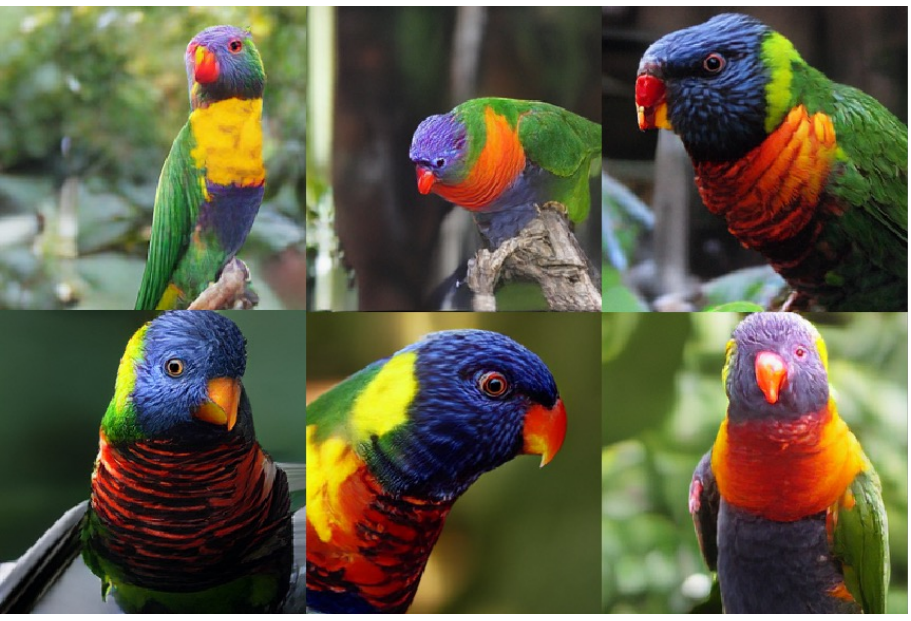}
\caption{Generated samples from MTAR-L for ImageNet class ID 90 (lorikeet).}
\label{fig:mtar_lorikeet}
\end{figure}

\begin{figure}[h]
\centering
\includegraphics[width=0.95\linewidth]{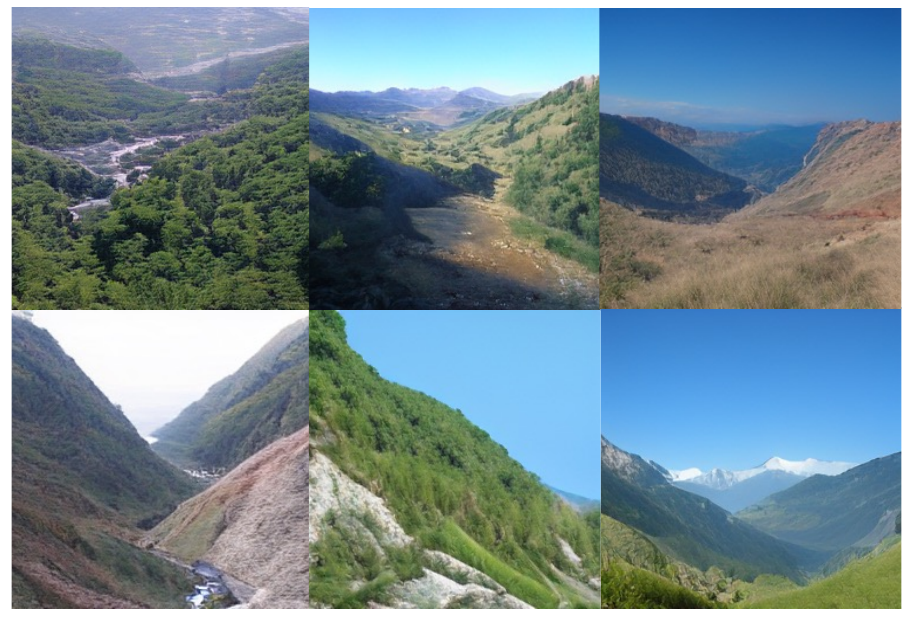}
\caption{Generated samples from MTAR-L for ImageNet class ID 979 (valley).}
\label{fig:mtar_valley}
\end{figure}

\section{Comparison with LlamaGen}
\label{sec:E}

\begin{figure}[H]
\centering
\vspace{-0.18in}
\includegraphics[width=0.92\linewidth]{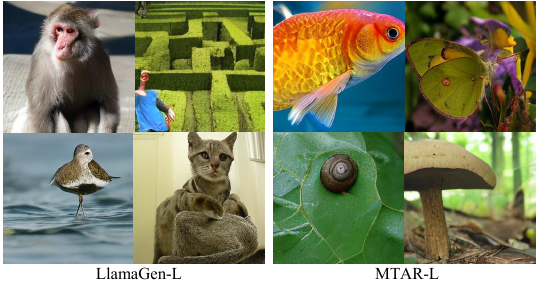}
\vspace{-0.17in}
\caption{Uncurated random samples from LlamaGen and MTAR.}
\vspace{-0.21in}
\label{fig:random_photo}
\end{figure}

\begin{figure*}[ht]
\centering
\includegraphics[width=0.92\linewidth]{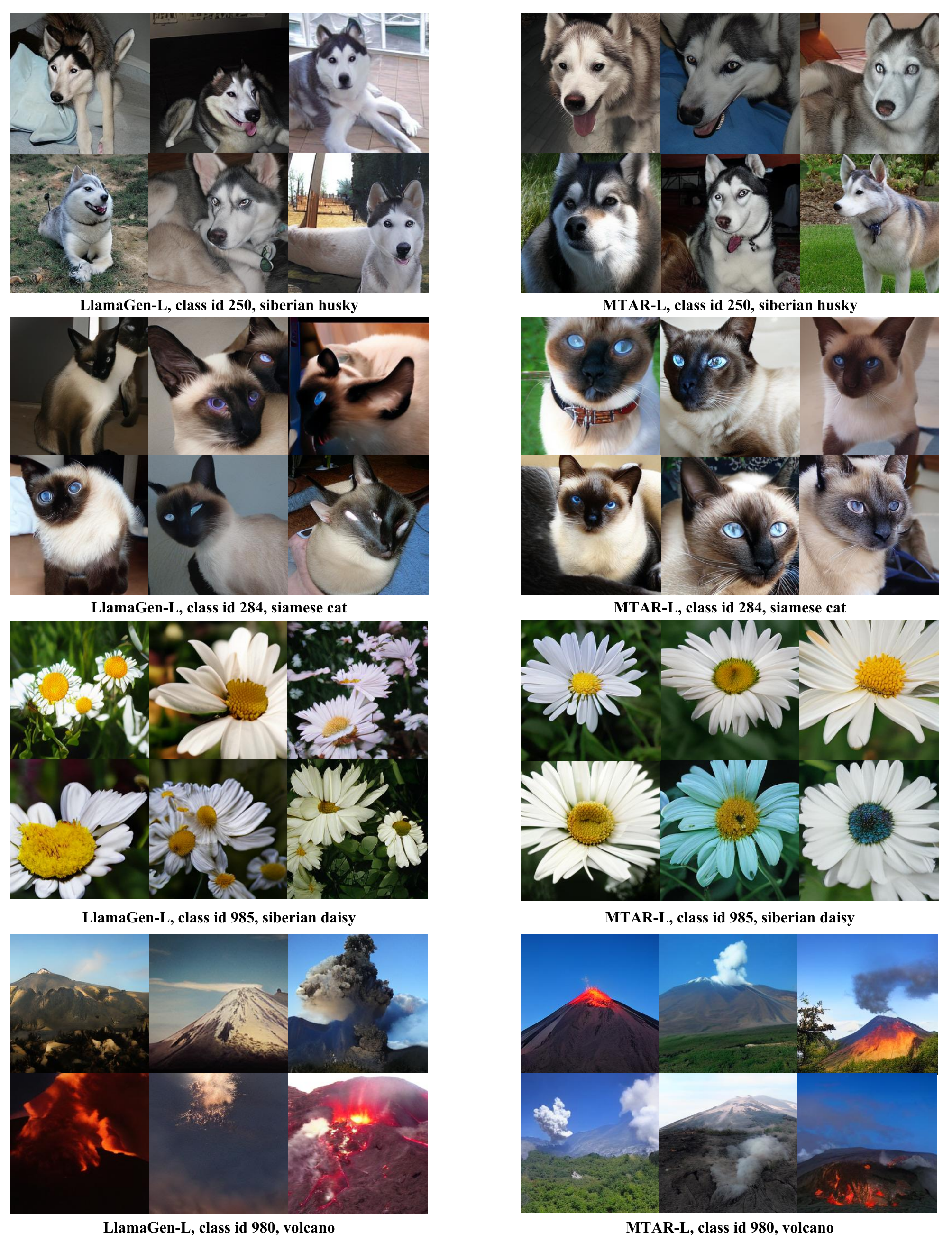}
\caption{Comparison of class-conditional samples produced by
LlamaGen-L and MTAR-L on ImageNet
$256\times256$.}
\label{fig:class_comparison_set1}
\end{figure*}

\begin{figure*}[ht]
\centering
\includegraphics[width=0.92\linewidth]{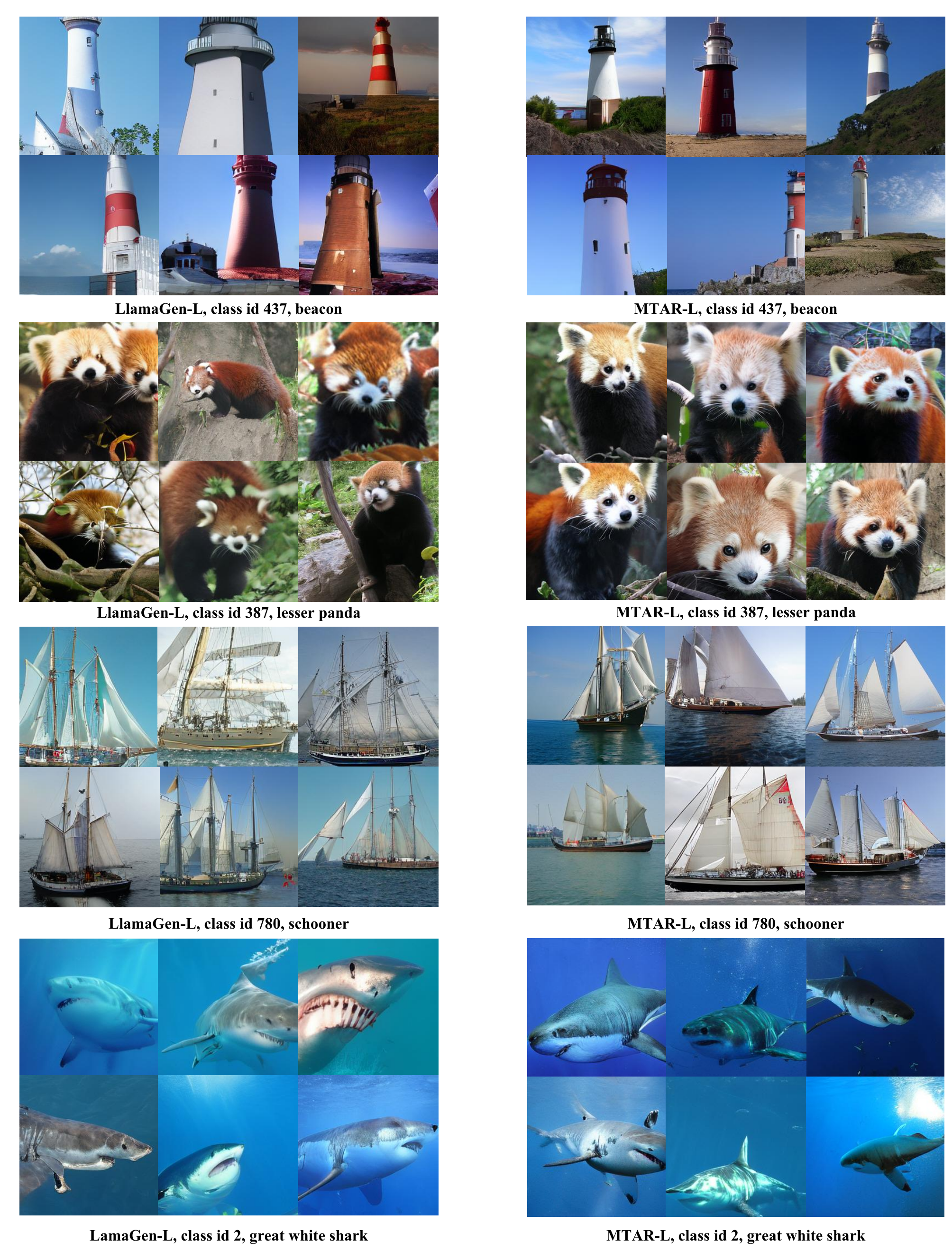}
\caption{Additional comparison of class-conditional samples produced by
LlamaGen-L and MTAR-L on ImageNet
$256\times256$.}
\label{fig:class_comparison_set2}
\end{figure*}

\end{document}